\documentclass[11pt,a4paper]{article}
\usepackage{times,latexsym}
\usepackage{url}
\usepackage[T1]{fontenc}

\usepackage[acceptedWithA]{tacl2018v2}
\usepackage{xspace,mfirstuc,tabulary}

\newif\iftaclinstructions
\taclinstructionsfalse 
\iftaclinstructions
\renewcommand{\confidential}{}
\renewcommand{\anonsubtext}{(No author info supplied here, for consistency with
TACL-submission anonymization requirements)}
\newcommand{\instr}
\fi

\iftaclpubformat 

\else

\fi

\usepackage{booktabs}
\usepackage{xspace}
\usepackage{tabularx}
\usepackage{xhfill}
\usepackage{xcolor}
\usepackage{multirow}

\newcommand\gold{\textsc{Gold}\xspace}

\definecolor{forestgreen}{HTML}{009B55}
\definecolor{sepia}{HTML}{671800}
\definecolor{midnightblue}{HTML}{006795}
\definecolor{orangered}{HTML}{E24C00}

\newcommand\NtN{\textsc{rnd2rnd}\xspace}
\newcommand\BtN{\textsc{bert2rnd}\xspace}
\newcommand\NtB{\textsc{rnd2bert}\xspace}
\newcommand\Bzero{\textsc{bertShare}\xspace}
\newcommand\Rzero{\textsc{robertaShare}\xspace}
\newcommand\BtB{\textsc{bert2bert}\xspace}
\newcommand\NtG{\textsc{rnd2gpt}\xspace}
\newcommand\Gone{\textsc{gpt}\xspace}
\newcommand\BtG{\textsc{bert2gpt}\xspace}
\newcommand\RtG{\textsc{roberta2gpt}\xspace}

\newcolumntype{Y}{>{\centering\arraybackslash}X}

\title{Leveraging Pre-trained Checkpoints for Sequence Generation Tasks}

\author{Sascha Rothe \\
  Google Research \\
  {\small \tt rothe@google.com} \\\And
  Shashi Narayan \\
  Google Research \\
  {\small \tt shashinarayan@google.com} \\\And
  Aliaksei Severyn \\
  Google Research \\
  {\small \tt severyn@google.com}
}

\date{}

\begin{document}
\maketitle
\begin{abstract}
Unsupervised pre-training of large neural models has recently revolutionized Natural Language Processing.
By warm-starting from the publicly released checkpoints, NLP practitioners have pushed the state-of-the-art on multiple benchmarks while saving significant amounts of compute time.
So far the focus has been mainly on the Natural Language Understanding tasks.
In this paper, we demonstrate the efficacy of pre-trained checkpoints for Sequence Generation.
We developed a Transformer-based sequence-to-sequence model that is compatible with publicly available pre-trained BERT, GPT-2 and RoBERTa checkpoints and conducted an extensive empirical study on the utility of initializing our model, both encoder and decoder, with these checkpoints.
Our models result in new state-of-the-art results on Machine Translation, Text Summarization, Sentence Splitting, and Sentence Fusion.
\end{abstract}

\section{Introduction}
Unsupervised and self-supervised pre-training methods, such as ELMo \cite{elmo}, ULMFiT~\cite{ulmfit}, and more recently BERT~\cite{bert}, GPT and GPT-2~\cite{gpt,gpt2}, XLNet~\cite{xlnet_arxiv19} and RoBERTa \cite{roberta} have established a qualitatively new level of baseline performance for many widely used Natural Language Understanding (NLU) benchmarks including some of the most popular, like GLUE~\cite{glue} and SQuAD~\cite{squad}.

The most appealing part about this massive shift towards using large architectures pre-trained on large collections of texts is that the pre-trained checkpoints along with the inference code are made freely available.
This saves hundreds of TPU/GPU hours as warm-starting a model from a pre-trained checkpoint typically requires orders of magnitude fewer fine-tuning steps while delivering significant performance boosts.
More importantly, the ability to bootstrap from a state-of-the-art performing model such as BERT~\cite{bert} motivates the community to
greatly speed up the progress towards developing better and easily reusable NLU systems.

While we continue to observe an increasing number of papers building on top of BERT and/or GPT models reporting encouraging improvements on Glue, SQuAD, and other similar benchmarks, very little attention has been paid to using these pre-trained models to warm-start sequence-to-sequence (seq2seq) models.
It has been argued that the pre-training objective used by BERT is not well suited for tasks that require decoding texts, e.g., conditional text generation in machine translation and summarization \cite{xlnet_arxiv19}.
Nevertheless, it remains unclear to what extent employing such large models pre-trained on large collections of text
can be beneficial to warm-start sequence-to-sequence generation models.

In this paper, we have developed a Transformer-based sequence-to-sequence model that is compatible with publicly available pre-trained BERT, GPT-2 and RoBERTa checkpoints. We aim to provide an empirical answer to the following research question: \textit{what is the best way to leverage publicly available pre-trained checkpoints for warm-starting sequence generation models?} For example,
one could imagine using BERT checkpoint to initialize the encoder for better input understanding and choosing GPT-2 model as the decoder for better text generation.
One of the main contributions of this paper is that we rigorously experiment with a large number of different settings to combine BERT, GPT and RoBERTa pre-trained checkpoints to initialize our Transformer-based model. We report results on three canonical conditional text generation tasks of increasing complexity: sentence-level fusion
\citep[DiscoFuse,][]{discofuse} and splitting \citep[WikiSplit,][]{wikisplit}), WMT14 En$\leftrightarrow$De machine translation using most common eval sets: newstest2014 and newstest2016, and abstractive summarization using three datasets: Gigaword \cite{gigaword}, CNN and DailyMail \cite{hermann-nips15} and BBC extreme~\cite{narayan-etal-2018-xsum}.

Our models report significant improvements over randomly initialized models demonstrating the benefit of leveraging unsupervised pre-trained models. More importantly, this simple strategy results in new state-of-the-art results on Machine Translation, Text Summarization, Sentence Splitting, and Sentence Fusion. Our results also demonstrate that a pre-trained encoder is an essential component for sequence generation tasks and often these tasks benefit from sharing the weights between the encoder and the decoder. Overall, we have run over 300 experiments spending thousands of TPU v3 hours to better accommodate the language modeling and understanding capabilities of these pre-trained models for text generation. We believe that NLP researchers and  practitioners will derive actionable insights from our findings when tackling various seq2seq tasks.

The code to query our models and predictions on various benchmarks will be available at \href{https://github.com/google-research/google-research/tree/master/bertseq2seq}{https://github.com/google-research/google-re}
\href{https://github.com/google-research/google-research/tree/master/bertseq2seq}{search/tree/master/bertseq2seq}.

\section{Models and Pre-trained Checkpoints}
\label{sec:models}

BERT was primarily developed for encoding text representations for NLU tasks (encoder-only architecture), whereas GPT-2 \cite{gpt2}, as a decoder-only architecture for language modeling.
Our model uses a seq2seq architecture with encoder and decoder both composed of Transformer layers \cite{transformer}.
For the encoder, we inherit the BERT Transformer layer implementations \cite{bert}, which differs slightly from the canonical Transformer layer \cite{transformer}; BERT uses a \textsc{gelu} activation \cite{gelu} rather than the standard \textsc{relu}.
If not stated otherwise, the implementation of the decoder layers are also identical to the BERT implementation with two adjustments.
First the self-attention mechanism is masked to look only at the left context.
Secondly, we add an encoder-decoder attention mechanism.
Note, that if the model was randomly initialized, we found no difference between a BERT compatible decoder and a GPT-2 compatible decoder.

Most of the models use the base checkpoint and therefore have 12 layers, a hidden size of 768, filter size of 3072, and 12 attention heads.
We chose the best-performing model and also collect numbers using larger pre-trained checkpoints.
These models have 24 layers, a hidden size of 1024, filter size of 4096, and 16 attention heads.

All models were fine-tuned on the target task using Adam with a learning rate of 0.05. We used a linear learning rate warmup with 40k steps, normalization by the square root of the hidden size, and a square root decay. We did not perform any tuning of these hyperparameters (except for \S\ref{sec:ablation}). The batch size and the number of training steps will be reported for each task individually.

\noindent \textbf{BERT Checkpoints.}
We tokenize our text using the WordPiece \cite{wordpiece} to match the BERT pre-trained vocabulary.
Depending on the experiment, we use one of the following publicly available checkpoints: BERT-Base Cased, BERT-Base Uncased, BERT-Base Multilingual Cased \cite{bert}.\footnote{BERT checkpoints are available at \url{https://github.com/google-research/bert}.}
The first two checkpoints have a vocabulary size of around $\sim$30k wordpieces, whereas the multilingual checkpoint has a much larger vocabulary size of $\sim$110k. BERT also trains positional embeddings for up to 512 positions, which is the maximum input and output length in all experiments.

\noindent \textbf{GPT-2 Checkpoints.}
We tokenize our text using the SentencePieces \cite{sentencepiece} to match the GPT-2 pre-trained vocabulary.\footnote{GPT-2 checkpoints are available at \url{https://github.com/openai/gpt-2}.}
Note that, while the available checkpoint is frequently called 117M, which suggests the same number of parameters, we count 125M parameters in the checkpoint.
This is the smallest architecture they trained, and the number of layers, hidden size, and filter size are comparable to BERT-Base.
The model was trained mainly on English data but does contain some foreign language.
The vocabulary size is $\sim$50k. While GPT-2 has positional embeddings for up to 1024 position, we only use the first 512 to make the results comparable with BERT.

\noindent \textbf{RoBERTa Checkpoints.}
RoBERTa \cite{roberta} is trained using PyTorch, but we found that the learned parameters are fully compatible with the existing TensorFlow BERT architectures with some minor adjustments.\footnote{More specifically: a) the variable names have to be adjusted; b) the weight and bias variables of the attention mechanism have to be splitted into query, key, and values; c) all variables except the embedding matrices have to be transposed.} The vocabulary treatment in RoBERTa is compatible with the SentencePiece tokenization in GPT-2.\footnote{RoBERTa checkpoints are available at \url{https://github.com/pytorch/fairseq}.}
As the conceptual differences between BERT and RoBERTa are minor, we might use BERT as a hypernym to address both pretraining methods in this paper.

\begin{table}[]
\centering
\footnotesize
\begin{tabular}{l|cccc}
\toprule
       & total   & embed.     & init.       & random \\ \midrule
\NtN   & 221M    & 23M        & 0           & 221M   \\
\BtN   & 221M    & 23M        & 109M        & 112M   \\
\NtB   & 221M    & 23M        & 109M        & 26M   \\
\BtB   & 221M    & 23M        & 195M        & 26M    \\
\Bzero & 136M    & 23M        & 109M        & 26M    \\
\Rzero & 152M    & 39M        & 125M        & 26M \\
\Gone  & 125M    & 39M        & 125M        & 0      \\
\NtG   & 238M    & 39M        & 125M        & 114M   \\
\BtG   & 260M    & 62M        & 234M        & 26M    \\
\RtG   & 276M    & 78M        & 250M        & 26M \\
\bottomrule
\end{tabular}
\caption{The number of total trainable parameters, embedding parameters and  parameters initialized from the checkpoint vs. randomly. The BERT/GPT-2 embeddings have 23M/39M parameters. The encoder-decoder attention accounts for 26M parameters.}
\label{tab:models}
\end{table}

\section{Investigated Model Variants}

In this section, we describe several combinations of model initialization.
The number of total trainable parameters, the number of embedding parameters and the number of parameters initialized from the checkpoint vs. randomly are shown in Table~\ref{tab:models}.

\textbf{\NtN} A Transformer encoder-decoder architecture with all weights initialized randomly.

\textbf{\BtN} A BERT-initialized encoder paired with a randomly initialized decoder.
Encoder and decoder share the embedding matrix initialized from a checkpoint.

\textbf{\NtB} A randomly initialized encoder paired with a BERT-initialized decoder.
To perform autoregressive decoding, we mask the bidirectional self-attention mechanism of BERT to look only at the left context.

\textbf{\BtB} A BERT-initialized encoder paired with a BERT-initialized decoder. All weights are initialized from a public BERT checkpoint. The only variable that is initialized randomly is the encoder-decoder attention.

\textbf{\Bzero} Like \BtB, but the parameters between encoder and decoder are shared. This greatly reduces the memory footprint of the model (136M vs. 221M parameters). Additionally, we experimented with a layer-wise attention mechanism \cite{he2018layer}, but got nearly identical numbers on most tasks.

\textbf{\Rzero} Same as \Bzero, but the shared encoder and decoder are initialized with the public RoBERTa checkpoint.

\textbf{\Gone} A decoder-only architecture. We treat the input as a conditioning prefix of a language model. The decoder is warm-started with a public GPT-2 checkpoint. Similarly to \Bzero and \Rzero, the memory footprint of this model is smaller compared to an encoder-decoder setup (125M parameters).

\textbf{\NtG} A randomly initialized encoder paired with a GPT-2-compatible decoder. We warm-start the decoder and the embedding matrix with a public GPT-2 checkpoint.

\textbf{\BtG} A BERT-compatible encoder paired with a GPT-2-compatible decoder. We warm-start both sides with the two separate, BERT and GPT-2, public checkpoints. We use the BERT vocabulary for the input and the GPT-2 vocabulary for the output.

\textbf{\RtG} Same as \BtG, but we use a public RoBERTa checkpoint to warm-start the encoder.
RoBERTa was trained using the GPT-2 vocabulary so we can use it for input and output.
Note that while the vocabulary is shared, this model still has two embeddings matrices, one for the input and one for the output.

The pre-training objective in the BERT models learns to predict a masked token using the bidirectional representation of the input text \cite{bert,roberta}.
Our decoder, even when initialized with the BERT or RoBERTa checkpoints, always generates the output text in an autoregressive fashion as in Tranformers \cite{transformer} and GPT-2 \cite{gpt2}.

We performed the bulk of our experiments on the 12-layer checkpoints of BERT, GPT-2, and RoBERTa, assuming that the findings will also hold for the 24-layer checkpoints.
We chose \Bzero and \Rzero to also report numbers using the 24-layer public pre-trained checkpoints.
We also experimented with the \Gone setup with 24 layers and 345M parameters but as we did not achieve any better results we excluded this from the paper.

\section{Experiments and Results}

\subsection{Sentence Fusion}
Sentence Fusion is the problem of combining multiple sentences into a single coherent sentence.
We use the ``balanced Wikipedia'' portion of the DiscoFuse dataset \cite{discofuse} for our experiments with 4.5M fusion examples in the training set.
The evaluation set has 50k example.
Due to the size of this evaluation set, even small changes are statistically significant.
For this reason, we have solely chosen this dataset for additional experiments described at the end of the paper.

\begin{table}[]
\centering
\footnotesize
\begin{tabular}{l|cc|c|c}
\toprule
DiscoFuse & \multicolumn{2}{c|}{100\%} & 10\% & 1\%  \\
          & Exact  & SARI  & SARI & SARI  \\
\midrule
\cite{discofuse}   & 51.1          & 84.5          & --       & --         \\
\midrule
\multicolumn{5}{l}{\hspace{-.5em}\textbf{Initialized with the base checkpoint (12 layers)}} \\
\RtG & \textbf{65.6}   & \textbf{89.9} & \textbf{87.1} & 80.3    \\
\Rzero & 65.3          & 89.7          & 86.9     & \textbf{81.2}    \\
\BtB   & 63.9          & 89.3          & 86.1     & \textbf{81.2}   \\
\BtN   & 63.9          & 89.3          & 86.1     & 80.3    \\
\Bzero & 63.9          & 89.2          & 86.0     & 80.8    \\
\BtG   & 61.5          & 88.4          & 84.1     & 70.2   \\
\Gone  & 60.4          & 88.0          & 82.9     & 74.5   \\
\NtB   & 60.0          & 87.6          & 82.1     & 72.8    \\
\NtN   & 58.3          & 86.9          & 81.5     & 69.3    \\
\NtG   & 57.6          & 86.5          & 81.4     & 70.6   \\
\midrule
\multicolumn{5}{l}{\hspace{-.5em}\textbf{Initialized with the large checkpoint (24 layers)}} \\
\Rzero & \textbf{66.6} & \textbf{90.3} & \textbf{87.7} & \textbf{81.5}    \\
\Bzero & 65.3          & 89.9          & 86.6          & 81.4    \\
\bottomrule
\end{tabular}
\caption{Results of different models and initialization techniques on DiscoFuse and subsampled training sets. Blockwise sorted by SARI score on 100\% of the training set.}
\label{tab:discofuse}
\end{table}

Training was done for 300k steps with a global batch size of 256. The input and output are padded to a length of 128, which covers 100\% of the training, evaluation and test data. We report SARI \cite{xu-etal-2016-optimizing}\footnote{\label{sari_implementation}SARI
is a lexical similarity metric which compares the model's output to multiple references and the input in order to assess the model's ability to add, delete and keep an $n$-gram. It's implementation is available at:
\texttt{\href{https://github.com/tensorflow/tensor2tensor/blob/master/tensor2tensor/utils/sari_hook.py}{https://} \href{https://github.com/tensorflow/tensor2tensor/blob/master/tensor2tensor/utils/sari_hook.py}{github.com/tensorflow/tensor2tensor/blob/} \href{https://github.com/tensorflow/tensor2tensor/blob/master/tensor2tensor/utils/sari_hook.py}{master/tensor2tensor/utils/sari\_hook.py}}.}
and the exact match accuracy. The results can be seen in Table~\ref{tab:discofuse}.
Previous state-of-the-art results by \newcite{discofuse} used the vanilla transformer model by \newcite{transformer}, with only 7 layers.
All models with initialized encoders outperform the baseline by a large margin, with a SARI score of 89.3 compared to 86.9 (\BtN vs. \NtN).
To measure the effect on smaller training sets, we randomly subsample the training data down to 10\% and 1\%, i.e. 450k and 45k training examples, respectively.
First, we notice, that performance comparable to the baseline is achieved even when training on only 10\% of the training data (\NtN vs. \Rzero).
Secondly, when using only 1\% of the training data setups with fewer randomly initialized parameters (\BtB vs. \BtN) perform better.
The best performing 12 layer setup is \RtG with a SARI score of 89.9 only outperformed by 24 layer setup of \Rzero with a SARI score of 90.3.

\begin{table}[]
\centering
\footnotesize
\begin{tabular}{l|ccc}
\toprule
WikiSplit & Exact  & SARI   & BLEU \\
\midrule
\cite{wikisplit} & 14.3        & 61.5       & 76.4  \\
\midrule
\multicolumn{4}{l}{\hspace{-.5em}\textbf{Initialized with the base checkpoint (12 layers)}} \\
\Bzero & \textbf{16.3} & \textbf{63.5}   & \textbf{77.2}  \\
\Rzero & 16.1        & 63.4       & 77.1  \\
\BtB   & 15.6        & 63.2       & 77.0  \\
\RtG   & 15.1        & 63.2       & 76.8  \\
\BtN   & 15.9        & 63.1       & 76.9  \\
\BtG   & 14.6        & 62.4       & 76.5  \\
\NtB   & 15.2        & 61.8       & 76.5  \\
\NtN   & 14.6        & 61.7       & 76.3  \\
\NtG   & 14.2        & 61.3       & 76.2  \\
\Gone  & 14.2        & 61.1       & 75.8  \\
\midrule
\multicolumn{4}{l}{\hspace{-.5em}\textbf{Initialized with the large checkpoint (24 layers)}} \\
\Rzero & 16.4 & \textbf{63.8}   & \textbf{77.4}  \\
\Bzero & \textbf{16.6} & 63.7   & 77.3  \\
\bottomrule
\end{tabular}
\caption{Results of different models and initialization setups on WikiSplit. Blockwise sorted by SARI score.}
\label{tab:wikisplit}
\end{table}

\subsection{Split and Rephrase}
The reverse  task  of  sentence  fusion  is  the split-and-rephrase task, which requires rewriting a long sentence into two or more coherent short sentences \cite{narayan-etal-2017-split}.
We use the WikiSplit dataset \cite{wikisplit}, which consists of 1M examples of sentence splits extracted from the Wikipedia edit history, and follow the training/test split suggested by the authors.
Training was done for 300k steps with a global batch size of 256.
The input and output are padded to a length of 128, which covers 100\% of the training, evaluation and test data.
As in \newcite{wikisplit}, we report corpus-level BLEU\footnote{We use NLTK v3.2.2 with case sensitive scoring to estimate BLEU scores.}, the exact match accuracy, and SARI score.
Previous state-of-the-art results by \newcite{wikisplit} used a bi-directional LSTM with a copy mechanism \cite{aharoni-goldberg-2018-split}.
Analogous to the DiscoFuse task we observe that initializing the encoder improves the model the most (Table~\ref{tab:wikisplit}).
The shared encoder-decoder setup of \Bzero outperforms all other setups.
For the larger models with 24 layers, we observed a small over-fitting after 100k steps (\textasciitilde25 epochs), and therefore stop the training early.
\Bzero and \Rzero perform on par and both outperform their 12 layer counterpart.

\subsection{Machine Translation}
We test our setups on the most common benchmark in machine translation -- WMT 2014 English $\leftrightarrow$ German task -- using newstest2014 and newstest2016 eval sets. We use the same hyper-parameter settings as in the previous experiments. We limit the input and output lengths to 128 tokens each.
We used a global batch size of 256 and train for 30 epochs. 
Decoding was done with the beam size of 4 and the default value for the sentence length penalty is set to $\alpha=0.6$.
We report uncased BLEU-4 scores.\footnote{We use a script from the Tensorflow Official Transformer implementation
\texttt{\href{https://github.com/tensorflow/models/tree/master/official/nlp/transformer}{https://github.com/tensorflow/models/tree}
\href{https://github.com/tensorflow/models/tree/master/official/nlp/transformer}{master/official/nlp/transformer}}.
Note that, differently from the
\texttt{\href{https://github.com/tensorflow/tensor2tensor/blob/master/tensor2tensor/utils/get_ende_bleu.sh}{tensor2tensor/utils/} \href{https://github.com/tensorflow/tensor2tensor/blob/master/tensor2tensor/utils/get_ende_bleu.sh}{get\_ende\_bleu.sh}}
used by \newcite{transformer}, this script does not split noun compounds, but we normalize utf-8 quotes to ascii quotes as we noted that our pre-processed training set contains only ascii quotes.}

\begin{table*}[tb]
\centering
\footnotesize
\begin{tabular}{l|cc|cc}
\toprule
& \multicolumn{2}{c|}{newstest2014} & \multicolumn{2}{c}{newstest2016} \\
& En$\rightarrow$De & De$\rightarrow$En & En$\rightarrow$De & De$\rightarrow$En\\
\midrule
\cite{transformer} & 27.3 & -- & -- & -- \\
Transformer (ours) & 28.1 & 31.4 & 33.5 & 37.9 \\
KERMIT~\cite{kermit} & 28.7 & 31.4 & -- & -- \\
\cite{transformer-nmt} & 29.2 & -- & -- & -- \\
\cite{backtranslation}* & \textbf{35.0} (33.8) & -- & -- & -- \\
\midrule
\multicolumn{5}{l}{\hspace{-.5em}\textbf{Initialized with public checkpoints (12 layers) and vocabulary}} \\
Transformer (ours) & 23.7 & 26.6 & 31.6 & 35.8 \\
\NtN & 26.0 & 29.1  & 32.4 & 36.7 \\
\BtN & \textbf{30.1} & \textbf{32.7}  & 34.4 & \textbf{39.6} \\
\NtB & 27.2 & 30.4  & 33.2 & 37.5 \\
\BtB & \textbf{30.1} & \textbf{32.7}  & \textbf{34.6} & 39.3 \\
\Bzero & 29.6 & 32.6  & 34.4 & \textbf{39.6} \\
\Gone & \textcolor{gray}{16.4} & 21.5 & \textcolor{gray}{22.4} & 27.7 \\
\NtG & \textcolor{gray}{19.6} & 23.2  & \textcolor{gray}{24.2} & 28.5 \\
\BtG & \textcolor{gray}{23.2} & 31.4  & \textcolor{gray}{28.1} & 37.0 \\
\midrule
\multicolumn{5}{l}{\hspace{-.5em}\textbf{Initialized with a custom BERT checkpoint (12 layers) and vocabulary}} \\
\BtN & \textbf{30.6} & 33.5  & 35.1 & \textbf{40.2} \\
\Bzero & 30.5 & \textbf{33.6}  & \textbf{35.5} & 40.1 \\
\midrule
\multicolumn{5}{l}{\hspace{-.5em}\textbf{Initialized with a custom BERT checkpoint (24 layers) and vocabulary}} \\
\BtN & \textbf{31.7} & \textbf{34.2}  & \textbf{35.6} & \textbf{41.1} \\
\Bzero & 30.5 & 33.8 & 35.4 & 40.9 \\
\bottomrule
\end{tabular}
\caption{Uncased BLEU-4 scores on WMT14 English $\leftrightarrow$ German newstest2014 and newstest2016 test sets. Models in the middle section use the 110k wordpiece vocabulary that comes with the multilingual BERT checkpoint. In the bottom section, we use the native 32k wordpiece vocabulary extracted from WMT14 train set and a BERT checkpoint pre-trained only on English and German subset of Wikipedia. * leveraging a large number of additional parallel sentence pairs obtained with back-translation; we include this score as a reference to the highest achieved result on newstest2014. The GPT-2 results for En$\rightarrow$De (where the GPT-2 initialized decoder is used to decode targets in De) are grayed out as they are apriori penalizing for GPT-2 which was only pretrained on En texts.}\label{tab:nmt-results}
\end{table*}

In Table~\ref{tab:nmt-results}, we first report the baseline scores for the original Transformer model \newcite{transformer} and our Transformer implementation\footnote{We use Transformer layers from the official BERT implementation which have small differences from ~\cite{transformer}.} with the same hyper-parameters. In both cases, we use the encoder and decoder with 6 layers and the 32k wordpiece vocabulary extracted from the WMT14 training set. Our implementation obtains slightly higher scores than the original implementation.

The middle section of Table~\ref{tab:nmt-results} reports the results for various initialization schema using BERT and GPT-2 pre-trained checkpoints. Note that here all models have encoders and decoders with 12 layers. For BERT models, we use the BERT-Base Multilingual Cased checkpoint to initialize the encoder or the decoder or both, as the task involves one non-English language. This checkpoint has been pre-trained on 108 languages using a multilingual Wikipedia dump with a vocabulary of 110k wordpieces.
First, we observe that initializing the model with the BERT checkpoint is most beneficial on the encoder side; our observation is in line with \newcite{xlnet_arxiv19}.
Furthermore, models initialized with the BERT checkpoint receive a significant boost: \BtN compared to the no-initialization \NtN setup scores higher by +4 points on En$\rightarrow$De and +3.6 points on De$\rightarrow$En on newstest2014.
Contrary to the WikiSplit and DiscoFuse task, sharing the encoder and decoder variables did not give an additional boost.
This is most likely because a) model capacity is an important factor in MT and b) encoder and decoder have to deal with different grammar and vocabulary.

GPT-based models (\NtG, \Gone, and \BtG) do not perform nearly as well, especially when GPT is used as the decoder and the target language is German. This is because the GPT model comes with an English vocabulary and has been pre-trained mainly on English text. Hence, we report the scores for GPT in the En$\rightarrow$De setting in gray.

\noindent \textbf{Customized BERT checkpoint.} For this experiment we did not include RoBERTa, as the public checkpoint is available for English only. Instead, we train our own checkpoint.
We also observe that our implementation of the baseline Transformer, as well as \NtN setup which uses no initialization, perform weaker on newstest2014 compared to the Transformer baselines (with 6 layers and the 32k wordpiece vocabulary) we report in the top section of Table~\ref{tab:nmt-results}.
We conjecture that the differences might be due to the larger 110k wordpiece vocabulary trained to handle 104 languages from Wikipedia dump which is suboptimal for WMT14 data and leads to inferior results.
To verify this conjecture, we perform the following experiment: we use the 32k wordpiece vocabulary extracted from the WMT14 En $\leftrightarrow$ De training set (same as used in the top section of Table~\ref{tab:nmt-results}) and pre-train a BERT model on the English and German subset of the Wikipedia dump in the same way as the multilingual BERT checkpoint was obtained.
We initialize our best-performing setups, \BtN and \Bzero, with this checkpoint (the third block of Table~\ref{tab:nmt-results}).
This provides a further +0.5 (En $\leftrightarrow$ De) and +0.8 (De $\leftrightarrow$ En) BLEU improvements on newstest2014, and, +1.1 and +0.7 on newstest2016, yielding an overall very strong performance on the challenging WMT14 task.
Experiments with the larger models (the last block) show further improvements of up to +1.1 BLEU points.

\newcite{backtranslation} report better results when they augment the training set with a massive amount of back-translated sentence pairs. To the best of our knowledge, among the approaches that only leverage parallel data from WMT14, our results are state-of-the-art on both newstest2014 and newstest2016.

\subsection{Abstractive Summarization}

Document summarization is the task of producing a short version of a document while preserving its salient information content.
We evaluate our setups on three different summarization datasets of varying characteristics: Gigaword \cite{gigaword}, CNN and DailyMail \cite{hermann-nips15}, and BBC extreme \cite{narayan-etal-2018-xsum}.
The Gigaword dataset focuses on abstractive sentence summarization with a total of 3.8M sentence-summary training pairs. The other two datasets focus on single-document summarization: the CNN/DailyMail dataset consists of 287k document-summary pairs, whereas the BBC dataset consists of 204k document-summary pairs.
The CNN/DailyMail summaries are in the form of bullet-point story highlights and exhibit a high degree of extraction, requiring the models to learn to copy from the source documents. The BBC summaries, on the other hand, are extreme, in that the documents are summarized into single-sentence summaries. These summaries demonstrate a high level of abstractiveness, and generating them automatically requires document-level inference, abstraction, and paraphrasing.

\begin{table*}[t!]
\centering
\footnotesize
\begin{tabularx}{\textwidth}{l|ccc|YYY|YYY}
\toprule
& \multicolumn{3}{c|}{Gigaword} & \multicolumn{3}{c|}{CNN/Dailymail} & \multicolumn{3}{c}{BBC XSum} \\
& R-1 & R-2 & R-L & R-1 & R-2 & R-L & R-1 & R-2 & R-L \\ \midrule
Lead & -- & -- & -- & 39.60 & 17.70 & 36.20 & 16.30 & 1.61 & 11.95 \\
PtGen & -- & -- & -- & 39.53 & 17.28 &  36.38 & 29.70 & 9.21 & 23.24 \\
ConvS2S & 35.88 & 17.48 & 33.29 & -- & -- & -- & 31.89 &  11.54 &  25.75 \\
MMN & -- & -- & -- & -- & -- & -- & 32.00 & 12.10 & 26.00 \\
Bottom-Up & -- & -- & -- & 41.22 & 18.68 & 38.34 & -- & -- & -- \\ \midrule
MASS & \textit{38.73} & 19.71 & \textit{35.96} & -- & -- & -- & -- & -- & -- \\
TransLM & -- & -- & -- &  39.65 & 17.74 & 36.85 & -- & -- & -- \\
UniLM & -- & -- & -- & \textit{43.47} & \textit{20.30} & \textit{40.63} & -- & -- & --\\ \midrule
\multicolumn{10}{l}{\hspace{-.5em}\textbf{Initialized with the base checkpoint (12 layers)}} \\
\NtN & 36.94 & 18.71 & 34.45 & 35.77 & 14.00 & 32.96 & 30.90 & 10.23 & 24.24 \\
\BtN & 37.71 & 19.26 & 35.26 & 38.74 & 17.76 & 35.95 & 38.42 & 15.83 & 30.80 \\
\NtB & 37.01 & 18.91 & 34.51 & 36.65 & 15.55 & 33.97 & 32.44 & 11.52 & 25.65 \\
\BtB & 38.01 & 19.68 & 35.58 & 39.02 & 17.84 & 36.29 & 37.53 & 15.24 & 30.05\\
\Bzero & 38.13 & \textbf{\textit{19.81}} & \textbf{35.62} & 39.09 & 18.10 & 36.33 & 38.52 & 16.12 & 31.13 \\
\Rzero & \textbf{38.21} & 19.70 & 35.44 & \textbf{40.10} & \textbf{18.95} & \textbf{37.39} & \textbf{39.87} & \textbf{17.50} & \textbf{32.37} \\
\Gone & 36.04 & 18.44 & 33.67 & 37.26 & 15.83 & 34.47 & 22.21 & 4.89	& 16.69 \\
\NtG & 36.21 & 18.39 & 33.83 & 32.08 & 8.81 & 29.03 & 28.48 & 8.77 & 22.30 \\
\BtG & 36.77 & 18.23 & 34.24 & 25.20	& 4.96	& 22.99 & 27.79 & 8.37 & 21.91 \\
\RtG & 37.94 & 19.21 & 35.42 & 36.35 & 14.72 & 33.79 & 19.91 & 5.20 & 15.88 \\
\midrule
\multicolumn{10}{l}{\hspace{-.5em}\textbf{Initialized with the large checkpoint (24 layers)}} \\
\Bzero & 38.35 & 19.80 & 35.66 & 39.83 & 17.69 & 37.01 & 38.93 & 16.35 & 31.52 \\
\Rzero & \textbf{38.62} & 19.78 & \textbf{35.94} & \textbf{40.31} & 18.91 & \textbf{37.62} & \textbf{\textit{41.45}} & \textbf{\textit{18.79}} & \textbf{\textit{33.90}} \\
\midrule
\end{tabularx}
\caption{Summarization results of different models and their initialization setups.
We compare our setups (the bottom block) against both non-pre-trained (the top block) and pre-trained (the middle block) models on various datasets: the Lead baseline, PtGen \cite{see-acl17}, ConvS2S \cite{convseq2seq}, MMN \cite{Kim:2018:arXiv}, Bottom-Up \cite{gehrmann-emnlp18},
MASS \cite{mass_icml19}, TransLM \cite{DBLP:journals/corr/abs-1905-08836} and UniLM \cite{unilm_arxiv19}. The Lead results for the CNN/DailyMail dataset is taken from \newcite{narayan-etal-2018-ranking}, whereas, Lead, PtGen and ConvS2S results on the BBC dataset are taken from \newcite{narayan-etal-2018-xsum}. Our best results are \textbf{boldfaced} and the best results on the datasets are \textit{italicized}.}
\label{tab:summarization}
\end{table*}

In all three cases, we did not anonymize entities. We worked on the original cased versions of CNN/DailyMail and BBC datasets. For Gigaword we used the lowercased version to match the requirements of the publicly available lowercased test set. During training, the input documents were truncated to 512 tokens for CNN/DailyMail and BBC, and to 128 tokens for Gigaword. Similarly, the length of the summaries was limited to 128 tokens for CNN/DailyMail, 64 for BBC, and 32 for Gigaword. We used a global batch size of 128 document-summary pairs for CNN/DailyMail and BBC, and 256 document-summary pairs for Gigaword. We adapted to different number of training steps depending on the training data sizes. Models were trained for 500k, 300k and 200k steps for the Gigaword, CNN/DailyMail and BBC summarization datasets respectively. In all cases, we used the standard publicly available test sets; these consists of 1951 instances for Gigaword, 11490 for CNN/DailyMail and 11334 for BBC. We report on the ROUGE $F_1$ scores \cite{rouge}; in particular, we report on ROUGE-1 and ROUGE-2 for informativeness and ROUGE-L for fluency in Table~\ref{tab:summarization}.

\noindent \textbf{Document understanding.}
All BERT encoder based setups (i.e., \BtN, \Bzero, \Rzero, and \BtB) outperform the baseline \NtN by a large margin. The improvements of the \NtB setup, where only the decoder is initialized, are narrow. These results overall validate the significance of document representation in the encoder-decoder framework for summarization. On the BBC extreme summarization in particular, these four models achieve on average +6.85 point improvement in ROUGE-L compared to the \NtN setup. Our results demonstrate that the models with better document representations are better in generating extreme summaries that require document-level inference and abstraction. For the extractive highlights in the CNN/DailyMail dataset, these models show an improvement of +3.53 ROUGE-L points over the \NtN baseline. For Gigaword, where the input is a single sentence, the improvements are minimal (average of +1.02 ROUGE-L points).
The \Bzero setup with shared encoder and decoder parameters achieves better performance than \BtB on all three datasets. The gains are larger on the BBC dataset than on the Gigaword and CNN/DailyMail datasets. This is probably because the BBC summary sentences follow a distribution that is similar to that of the sentences in the document, whereas this is not necessarily the case for the Gigaword headlines and the CNN/DailyMail bullet-point highlights.
\Rzero performs superior to \Bzero on the CNN/DailyMail and BBC datasets.
\Rzero performs competitively to \Bzero on the Gigaword dataset where the task is to summarize sentences.

\noindent \textbf{Summarization with GPT checkpoints.}
\Gone (decoder-only) performs better than \NtG, \BtG or \RtG (encoder-decoder models) by a large margin for generating CNN/DailyMail extracts, but poorer for generating BBC abstracts. The encoder-decoder architecture where the input document is modeled separately is better equipped for document-level abstraction than the decoder-only architectures where the input document is a conditioning prefix of a language model. Initialization with different checkpoints, e.g., encoder with BERT and decoder with GPT in \BtG, is not effective for document summarization; \BtG and \RtG are inferior to \NtG on the BBC dataset and \BtG, to \NtG on the CNN/DailyMail dataset. However, this is not the case with the Gigaword dataset, which has 3.8M training instances; \BtG and \RtG perform better than \NtG.

\Rzero performs the best and is on par with the current state-of-the-art MASS model \cite{mass_icml19} on the Gigaword dataset. The MASS model has an advantage of pre-training encoder-decoder attention from scratch, our proposed models use the publicly available pre-trained checkpoints and only fine-tune on the target task. It is not obvious how the masked seq2seq pre-training objective for sentence generation in the MASS model will be beneficial for tasks like document summarization. Our proposed models provide a generic alternative and can be easily adapted to various text generation tasks. The \Rzero setup sets a new state-of-the-art, outperforming all existing baselines by a large margin on the BBC extreme summarization task. The best model on the CNN/DailyMail dataset outperforms the Pointer Generator network \cite{see-acl17} and the pre-trained single-decoder model with TransformerLM \cite{DBLP:journals/corr/abs-1905-08836}. Our model, however, lags behind the Bottom-Up system \cite{gehrmann-emnlp18} with a task-specific module for content selection along with the copy mechanism \cite{gu-EtAl:2016} and the UniLM model \cite{unilm_arxiv19} with BERT-Large pre-trained for Bidirectional, unidirectional and seq2seq language modeling objectives. The UniLM model is also fine-tuned with an additional extractive summarization objective to predict relevant sentences in the document; this objective could be beneficial to generate the CNN/DailyMail extracts.

\section{Discussion on Ablation Studies}
\label{sec:ablation}

\paragraph{Combining Different Checkpoints.}
Combining BERT and GPT-2 into a single model (\BtG) did not work and often underperformed than a randomly initialized baseline.
This is presumable because the model has to learn two different vocabularies.
This argument is backed by the fact that for MT de$\rightarrow$en the \BtG setup performed well.
For this task the vocabulary setting is in favor of this particular task, meaning two vocabularies have to be learned anyways and the output is English, where GPT-2 was trained on.
Since RoBERTa and GPT-2 share the same vocabulary, combining both into a single model (\RtG) showed strong results on several tasks but did not outperform a setup where RoBERTa is used in the encoder and decoder.

\paragraph{Tuning GPT-2 Based Models.}
We were surprised that setups using the GPT-2 checkpoint performed relatively poorly given that it is trained as a language model on a large corpus; our intuition was that GPT-2 initialized decoders will be strong natural language generators.
To ensure that this was not due to an unfortunate choice of hyperparameters, we tuned the learning rate, the warmup steps, and the optimizer $\in$ \{Adam, Adafactor\} for the GPT-2 based setups (\NtG, \Gone, \BtG) on the DiscoFuse dataset.
Naturally, this gave us slightly higher numbers but not at a magnitude that would suggest a previously suboptimal setting.
Specifically, we got a SARI score of 88.8 compared to 88.4 for \BtG, 88.1 compared to 88.0 for \Gone and 87.7 compared to 86.5 for \NtG.

\paragraph{Initializing only Embeddings.}
We want to investigate the impact of the non-contextualized BERT and GPT-2 embeddings.
This means we are initializing the transformer model with only the embedding matrices.
The advantage of this setup would be that we could freely choose the model architecture and size and adapt it to a specific task.
We found almost no improvement over the fully randomly initialized model \NtN.
Concretely, we compute a SARI score of 87.1 using the BERT embeddings and 87.0 using the GPT-2 embeddings, compared to 86.9 of the \NtN baseline.
We observe slightly higher improvements of up to 2 percentage points when training on only 10\% of the training data.

\paragraph{Initializing only Layers.}
Contrary to the previous paragraph, we want to investigate the effect of initializing everything but the word embedding matrix.
The embedding matrix accounts for only 10-31\% of all learnable parameters and sometimes the vocabulary given from a public checkpoint might not be optimal for a certain task.
In these cases, it would be nice to redefine the vocabulary while still leveraging the checkpoint.
First, we remove the embeddings matrices from the warm-started variables and observe a drop of 1.7 points using the \Bzero setup and 11 points using the \Gone setup (Table \ref{tab:no_embeddings}).
The latter is probably due to the large vocab of the GPT-2 model which now remains random initialized.
We then train a new BPE model with 16k tokens using the DiscoFuse training data \cite{sentencepiece,sennrich2015neural}.
We observe almost no change on \Bzero, suggesting that the BERT vocabulary was already optimal for DiscoFuse.
\Gone however, showed a significant improvement using this much smaller vocabulary but is still behind the fully initialized setup.
Finally, we experimented with a more sensitive way of training the model, meaning that we fix all warm-started variables for 100k steps.
During this pre-training phase, we only train the new word embeddings.
After the pre-training, we fine-tune the entire model for another 300k steps.
This training scheme resulted in an improvement of 0.5 for the \Bzero setup, but overall the number is still way behind the fully initialized setup.
For \Gone, this training scheme did not result in a satisfying training curve.

\begin{table}[t!]
\centering
\footnotesize
\begin{tabular}{l|c|c}
\toprule
 & \Bzero & \Gone \\
\midrule
DiscoFuse                      & 89.3 & 88.0 \\
- embeddings from checkpoint   & 87.5 & 77.0 \\
+ task specific SentencePieces & 87.5 & 84.2 \\
+ pre-training SentencePieces   & 88.0 & 69.7 \\
\bottomrule
\end{tabular}
\caption{SARI scores on the Discofuse dataset when experimenting with different embedding setups. Each row also includes the setups of all previous rows.}
\label{tab:no_embeddings}
\end{table}

\paragraph{Initializing a Subset of Layers.}

Motivated by the results of using 24 layers, we want to investigate if only a subset of these 24 layers can be used.
To account for the larger hidden layer size (1024 vs. 768) and filter size (4096 vs. 3072) we limit ourselves to using only 10 layers and the embedding matrix of this model.
This model still has more parameters then the base model (324M vs. 221M for \BtB, 198M vs. 136M for \Bzero) but can be trained with the same batch size, in a comparable amount of time (3 min/1000 iterations).
As an initial experiment, we used the first 10 layers out of the large BERT checkpoint to initialize the \Bzero setup.
This gave us a SARI score of 88.2 on DiscoFuse, compared to 89.3 of using the base checkpoint and compared to 87.0 of using the embeddings only (see ``Initializing only Embeddings'').
We then performed a hyperparameter search on the evaluation set using CMA-ES \cite{DBLP:journals/corr/Hansen16a} to find an optimal subset of layers to use.
The best setup used the following layers: 9, 10, 13-18, 23, 24; and achieved a SARI score of 89.1.
While this is a remarkable improvement over using the first 10 layers, this setup is still outperformed by the base BERT model.

\section{Analysis of Abstractive Summaries}
\label{sec:summaryanalysis}

Finally we present a qualitative analysis of these models for text generation. In particular, we focused on extreme summarization which assesses models ability to do document-level inference and abstraction. We evaluated summaries from randomly initialized model (\NtN) and from best performing models initialized with GPT checkpoints (\NtG), BERT checkpoints (\Bzero) and RoBERTa checkpoints (\Rzero). We also included \gold summaries in our evaluation. Results are presented in Table~\ref{tab:summaryanalysis}.

\begin{figure*}[t!]
  \center{\footnotesize
    \begin{tabular}{l p{13cm}}
    \toprule
    \NtN & The Queen has celebrated her 90th birthday with a message on social media about \textcolor{orangered}{her 90th birthday}. \\
    \NtG & The Queen has celebrated her 90th birthday with a birthday celebration in Buckingham Palace. \\
    \Bzero & The Queen has paid \textcolor{orangered}{tribute to the Queen} by sending a tweet saying she was ``unwittingly \textcolor{orangered}{unwittingly unwittingly}.\\
    \Rzero & The Queen has sent a twitter message for her 90th birthday on twitter.\\
    \gold & The Queen has tweeted her thanks to people who sent her 90th birthday messages on social media.\\
    \midrule
    \NtN & Sir Bradley Wiggins says he is ``proud'' of being involved in the use of a banned steroid against Sir Bradley Wiggins. \\
    \NtG & Team Sky boss Sir Dave Brailsford says he is ``disappointed'' after team Sky agreed to change their contracts with \textcolor{orangered}{team Sky}. \\
    \Bzero & Team Sky boss Sir Dave Brailsford says he is ``\textcolor{orangered}{proud}'' of his team's handling of doping in cycling.\\
    \Rzero & Team Sky boss Dave Brailsford says he is ``not proud'' of his team's handling of allegations of wrongdoing in the sport.\\
    \gold & Team Sky boss Sir Dave Brailsford has said that his handling of the media following allegations against his team has made things a ``damn sight worse''.\\
    \midrule
    \NtN & A \textcolor{orangered}{19-year-old American singer has been shot} dead by police in San Francisco. \\
    \NtG & Police are investigating a shooting in the grounds of a music venue in Los Angeles.\\
    \Bzero & \textcolor{orangered}{US singer Chris Brown has been shot and wounded} at a gig in the US state of California.\\
    \Rzero & Five people have been shot dead in a shooting at a concert in California.\\
    \gold & Five people have been shot at a California nightclub while Chris Brown was performing.\\
    \midrule
    \NtN & A council has asked people \textcolor{orangered}{not to keep their toilets} in a bid to save money.\\
    \NtG & \textcolor{orangered}{People are being urged to use a ``ladies' toilet'' in Skye in Skye in Skye} by their own councillor.\\
    \Bzero & Complaints about the availability of public toilets on Skye and the isle of \textcolor{orangered}{Skye} is being investigated by highland council.\\
    \Rzero & Highland council has commissioned a review of public toilets and \textcolor{orangered}{public toilets} on Skye.\\
    \gold & Islanders on Skye have demanded greater availability of public toilets after complaints some visitors to the Isle are relieving themselves outside.\\
    \midrule
    \NtN & A man has been jailed for \textcolor{orangered}{six years} for posting offensive comments on Facebook about an \textcolor{orangered}{Aberdeen teenager} who was later found dead.\\
    \NtG & A man who \textcolor{orangered}{admitted killing his six-year-old friend in a disturbance in Aberdeen} has been jailed.\\
    \Bzero & A man who \textcolor{orangered}{admitted murdering} a toddler after posting offensive comments about him on Facebook has been jailed for \textcolor{orangered}{three years}.\\
    \Rzero & A man has been jailed \textcolor{orangered}{for three months} for posting ``vile'' abuse on Facebook about a missing toddler found dead in his \textcolor{orangered}{Aberdeenshire home}.\\
    \gold & A man who admitted posting offensive comments on Facebook about an Edinburgh boy beaten to death by his mother has been jailed for 12 months.\\
    \bottomrule
    \end{tabular}
  }
  \caption{Model generated and reference summaries used for human evaluation. Words in \textcolor{orangered}{orange} correspond to incorrect or repeated information.\label{fig:erroranalysis-xsum}}
\end{figure*}

\paragraph{Human Assessment of Summary Quality.}
The study was conducted on the Amazon Mechanical Turk platform using Best-Worst Scaling, a less labor-intensive alternative to paired comparisons \cite{louviere1991best,louviere2015best}. Our participants were presented with a document and summaries generated from two out of five systems (four models and gold summaries) and were asked to decide which summary was better than the other in order of informativeness --- \textit{does the summary capture important information in the document correctly and concisely?} --- and fluency --- \textit{is the summary written in well-formed English?} We randomly selected 40 documents from the XSum test set. We collected judgments from three different participants for each comparison. The order of summaries were randomized per document and the order of documents per participant. The score of a system was computed as the percentage of times it was chosen as best minus the percentage of times it was selected as worst. The scores range from -1 (worst) to 1 (best). See Figure~\ref{fig:erroranalysis-xsum} for few sample predictions that were used in our human evaluation.

\begin{table}[]
\centering
\footnotesize
\begin{tabularx}{0.45\textwidth}{l|c|c|c}
\toprule
& Length & Repetitions & Quality \\ \midrule
\NtN & 20.90 & 29.76 & -0.103\\
\NtG & 21.49 & \textbf{16.28} & -0.303\\
\Bzero & 20.71 & 27.03  & -0.097 \\
\Rzero & \textbf{21.70} & 28.68 & \textbf{0.153} \\ \hline
\gold & 24.61 & 4.66 & 0.347 \\
\midrule
\end{tabularx}
\caption{Qualitative and human evaluations of BBC extreme summaries. The lowest numbers for repetitions and the highest numbers for quality are bold faced. See the text for details.}
\label{tab:summaryanalysis}
\end{table}

Our participants found the \Rzero summaries to be the best in terms of their overall quality; the \Bzero summaries ranked second after \Rzero. We further carried out pairwise comparisons between all models to assess whether system differences are statistically significant.\footnote{One-way ANOVA with posthoc Tukey HSD tests; $p < 0.01$.} We did not observe significant differences between \NtN and \NtG, \NtN and \Bzero, and, \Rzero and \gold. All other differences were statistically significant.

\paragraph{Summary Lengths and Repetitions.}

All models generated summaries of comparable lengths; the average length of summaries is 20.90 for \NtN, 21.49 for \NtG, 20.71 for \Bzero and 21.70 for \Rzero. \Rzero produced summaries were closest to the \gold summaries in terms of length (21.70 vs 24.61).

Finally, we estimated the percentage of summaries with at least one repetition of rare or content words. We discarded the 500 most common words from the model generated and reference summaries, the rests were considered as rare or content words. \Bzero and \Rzero summaries improve over the \NtN summaries, but have  more repetitions than the \NtG summaries. See examples in Figure~\ref{fig:erroranalysis-xsum} for redundant repeated spans marked in \textcolor{orangered}{orange}.

Overall, \Bzero and \Rzero summaries are unequivocally better than \NtG summaries in terms of both automatic evaluations (assessing ROUGE) and human evaluations (assessing summary quality); there are still room for improvements in these models \cite{unilm_arxiv19,mass_icml19,bart}.

\section{Related Work}

\textbf{Representation learning.}
Starting around 2013, word embeddings like word2vec \cite{word2vec} or GloVe \cite{glove} became popular as they were easy to train in an unsupervised fashion on raw text and they improved several downstream tasks when used as features.
These word embeddings are invariant to the context the word is in.
There has been work to contextualize these embeddings, mainly to account for synonyms (e.g. \cite{huang2012improving,rothe2015autoextend}) before, but only in 2018 did training of the contextualized embeddings using large deep neural networks and an unsupervised training scheme become popular.

While ELMo \cite{elmo} and ULMFiT \cite{ulmfit} are based on LSTMs \cite{lstm}, BERT and GPT are based on the transformer architecture \cite{transformer}.
This architecture outperforms LSTMs on several NLP tasks and we therefore concentrated on these two pre-trained models.
The contextualized embedding for each input token is given by the corresponding output of the last encoder layer.

\textbf{Pre-training models.}
One can also see these models as pre-trained models \cite{DBLP:journals/corr/DaiL15a}, which are then fine-tuned for a downstream task.
This is the conceptual view we adopted for this paper.
Why unsupervised pre-training helps deep learning was investigated by \newcite{erhan2010does}.
While the unsupervised pre-training strategies are different from those used in our paper, we expect the findings to still hold.
They show that unsupervised pre-training is not simply a way of getting a good initial
marginal distribution, that classical regularization techniques cannot achieve the same performance as unsupervised pre-training, and that the effect of unsupervised pre-training
does not go away with more training data.
An extensive study of pre-training was done by \newcite{wang-etal-2019-tell}.
This study compares single sentence classification, sentence pair classification, sequence to sequence and language modeling tasks for pre-training and measures the effect on GLUE.
The primary results support the use of language modeling.
\newcite{DBLP:journals/corr/abs-1903-05987} explore whether it is preferable to fine-tune the entire model on a specific task or to use the learned representations as features, i.e. freezing the pre-trained model.
Their results suggest that the relative performance of fine-tuning vs. feature extraction depends on the similarity between the pre-training and the target tasks.
\newcite{wang2019tune} propose a combination of both, where first the model is trained with the BERT parameters being frozen and then the entire model is fine-tuned.
This is the training scheme we used in ``Initializing only Layers'' study.

\textbf{Pre-training for sequence generation.}
Pre-training for seq2seq learning was first done by \newcite{DBLP:journals/corr/RamachandranLL16}.
They used a language model to pre-train the encoder and decoder of an RNN seq2seq model.
Their method improved BLEU scores on newstest2014 by ~3 points and ROUGE-L on CNN/Dailymail also by 3 points.
However their BLEU score of 24.7 on newstest2014 En$\rightarrow$De, compared to 30.6 in this work, and 29.4 ROUGE-L on CNN/Dailymail, compared to 36.33 also show the superiority of the transformer model as well as the masked language model objective of BERT.
MASS \cite{mass_icml19} is a BERT-inspired method of pre-training sequence to sequence models.
One advantage of this method is that, in contrast to our setups (except for \Gone), the encoder-decoder attention mechanism is also pre-trained.
The downside of this approach is that the pre-trained model is task-specific and not as general as BERT or GPT-2.
UniLM \cite{unilm_arxiv19} also unifies bidirectional, unidirectional, and sequence to sequence
language modeling.
At the time of writing, no public checkpoint was available to us.
We compare our work with their results in Table \ref{tab:summarization}.
To overcome the issue that the encoder-decoder attention is not pre-trained,
\newcite{DBLP:journals/corr/abs-1905-08836} pre-trained a single transformer language model that encodes the source and generates the target.
This setup matches our \Gone setup.
\newcite{DBLP:journals/corr/abs-1901-07291} pre-train their model using casual language modeling (like GPT), masked language modeling (like BERT) and a third new objective called translation language modeling to improve cross-lingual pre-training.

\textbf{Leveraging public checkpoints.}
BERT has been used for various NLP tasks, such as question answering on the SQuAD dataset \cite{squad}.
It also achieved new state-of-the-art results on the GLUE benchmark \cite{glue} and grounded commonsense inference \citep[SWAG,][]{swag}.
All of these tasks are a form of classification or regression.
\newcite{DBLP:journals/corr/abs-1903-10318} fine-tuned BERT for Extractive Summarization.

An analysis of different layers of the BERT model was performed by \cite{DBLP:journals/corr/abs-1905-05950}.
They found that the classical NLP pipeline appears in the expected sequence.
In the context of our experiments in ``Initializing a Subset of Layers'', this would mean that the DiscoFuse task profits the most from pre-trained information about POS, constituents, dependencies and semantic roles.
A similar study by \newcite{jawahar:hal-02131630} found that BERT captures phrase-level information in the lower layers and linguistic information in intermediate layers, with surface features at the bottom, syntactic features in the middle and semantic features at the top.

GPT was also evaluated on natural language inference tasks.
In the extended version of GPT-2, the model was evaluated on more general natural language processing tasks, like machine translation, reading comprehension, summarization, and language modeling.
GPT-2 achieved new state-of-the-art results on several language modeling datasets.
On the other tasks, GPT-2 outperformed some unsupervised baselines but is still far behind supervised or task-specific approaches.

After we performed the majority of our experiments, XLNet \cite{xlnet_arxiv19}, an autoregressive pre-training method based on Transformer XL \cite{dai2019transformer} was released.
XLNet achieved new state-of-the-art results on several NLP task.
We leave the experiments with their public checkpoint for future work.

\section{Conclusion}

We performed an extensive study on leveraging pre-trained checkpoints for sequence generation.
Our findings show, that a pre-trained encoder is an essential part.
Most tasks also profit from sharing the weights between the encoder and the decoder, which additionally decreases the memory footprint.
While combing BERT and GPT-2 into a single model often underperformed a randomly initialized baseline, combining RoBERTa and GPT-2 achieves strong results and shows the importance of sharing the vocabulary.
Training a language specific BERT model also improves performance over using the multilingual version.

\section*{Acknowledgments}

We thank the reviewers and the action editor for  their feedback. We would like to thank Ryan McDonald, Joshua Maynez, and Bernd Bohnet for useful discussions.

\bibliography{TextGen}
\bibliographystyle{acl_natbib}

\end{document}